\RequirePackage{xcolor}  
\documentclass{article}

\usepackage{times}
\usepackage{graphicx} 
\usepackage{subfigure} 

\usepackage{natbib}

\usepackage{algorithm}
\usepackage{algorithmic}

\usepackage{hyperref}

\usepackage{amsmath}
\usepackage{amsfonts}
\usepackage{amssymb}

\usepackage{pgf}            

\usepackage[accepted]{icml2017}

\icmltitlerunning{Convolutional Networks for Spherical Data}

\begin{document} 

\twocolumn[
\icmltitle{Convolutional Networks for Spherical Signals}



\icmlsetsymbol{equal}{*}

\begin{icmlauthorlist}
\icmlauthor{Taco S. Cohen}{to}
\icmlauthor{Mario Geiger}{to}
\icmlauthor{Jonas Koehler}{to}
\icmlauthor{Max Welling}{to}
\end{icmlauthorlist}

\icmlaffiliation{to}{University of Amsterdam}

\icmlcorrespondingauthor{Taco S. Cohen}{taco.cohen@gmail.com}


\vskip 0.3in
]



\printAffiliationsAndNotice{}  

\newcommand{\Z}{\mathbb{Z}}
\newcommand{\R}{\mathbb{R}}
\newcommand{\C}{\mathbb{C}}

\newcommand{\GLR}[1]{\ensuremath{\operatorname{GL}(#1, \mathbb{R})}}
\newcommand{\SLR}[1]{\ensuremath{\operatorname{SL}(#1, \mathbb{R})}}
\newcommand{\SE}[1]{\ensuremath{\operatorname{SE}(#1)}}
\newcommand{\SO}[1]{\ensuremath{\operatorname{SO}(#1)}}
\newcommand{\Aff}[1]{\ensuremath{\operatorname{Aff}(#1, \mathbb{R})}}

\newcommand{\Tr}{\ensuremath{\operatorname{Tr}}}
\newcommand{\ind}{\ensuremath{\operatorname{ind}}}

\newcommand{\corr}{\ensuremath{\star}}
\newcommand{\conv}{\ensuremath{*}}

\newcommand\supp{\mathop{\rm supp}}

\begin{abstract}
    The success of convolutional networks in learning problems involving planar signals such as images is due to their ability to exploit the translation symmetry of the data distribution through weight sharing.
    Many areas of science and egineering deal with signals with other symmetries, such as rotation invariant data on the sphere.
    Examples include climate and weather science, astrophysics, and chemistry.

    In this paper we present spherical convolutional networks.
    These networks use convolutions on the sphere and rotation group, which results in rotational weight sharing and rotation equivariance.
    Using a synthetic spherical MNIST dataset, we show that spherical convolutional networks are very effective at dealing with rotationally invariant classification problems. 

\end{abstract}

\section{Introduction}

Today, neural network architectures are usually found by trial and error.
In an ideal world, we would have a few simple \emph{design principles} that would guide the design of neural networks for various tasks.
Although we are far from this goal, over the last few years, \emph{equivariance to symmetry transformations} has emerged as a powerful and widely applicable design principle.

Many learning problems have symmetries.
For instance, in image classification, the class label of an image remains the same when the image is shifted, so translations are symmetries of the label function.
Convolutional networks exploit this symmetry through weight sharing, thus achieving excellent statistical efficiency.
Key to this success is a property of convolutions called equivariance: a shift of the input leads to a shift of the output.
Because convolution layers are equivariant, they preserve the symmetry and can thus be stacked to create arbitrarily deep networks.

This idea, implicit in the classical ConvNet literature, can be generalized to other symmetries that the data may have.
A general theory of such group equivariant convolutional networks presented by \cite{Cohen2016, Cohen2017}.
By designing the network so that every layer is equivariant to transformations from the group of interest, the symmetry is preserved throughout the network and can be exploited through (generalized) convolutional weight sharing.



In this paper we show how a group equivariant network for the rotation group $\SO3$ acting on the 2-sphere $S^2$ can be constructed.
This is challenging for two reasons.
First, $S^2$ and $\SO3$ are continuous manifolds that do not admit symmetric sampling grids, which makes it somewhat tricky to rotate filters numerically (interpolation would be required).
Second, the group $\SO3$ is much bigger than the finite groups considered in earlier work on equivariant networks, making computational and memory efficiency paramount.
We address both of these challenges by using generalized Fast Fourier Transform (FFT) algorithms to compute the convolution.


The rest of the paper is structured as follows.
We first discuss related work on equivariant deep networks and generalized Fourier analysis.
Then, in section \ref{sec:s2so3conv} and \ref{sec:fft} we define spherical and $\SO3$ convolution and the generalized Fourier transform.
Section \ref{sec:s2cnn} puts the pieces together and describes the spherical convolutional network.
Finally, in section \ref{sec:experiments} we describe our experiments that verify the mathematical properties expected of the network, and demonstrate the effectiveness of the inductive bias of spherical ConvNets for rotation-invariant classification on a synthetic dataset of spherical MNIST digits.

\section{Related Work}
\label{sec:related-work}


A lot of recent work has focussed on exploiting symmetries for data-efficient deep learning.
Theoretical investigations of \cite{Anselmi2013} point out the potential for significant improvements in sample efficiency from deep networks that incorporate geometrical prior knowledge.
Successful implementations of equivariant neural networks include \cite{Worrall,Jacobsen2017a,Zhou2017,Ravanbakhsh2016,Dieleman2016,Cohen2016,Cohen2017}.
Group invariant scattering networks were explored by \cite{Sifre2013,Oyallon2015}.

\section{Spherical and \SO3-Convolution}
\label{sec:s2so3conv}

By analogy to a planar convolution, which is computed by taking inner products between a signal and a shifted filter, we define the spherical convolution in terms of the inner product between a spherical signal and a rotated spherical filter.
Both the signal $f$ and filter $\psi$ are considered to be vector-valued functions on the sphere, i.e. $f : S^2 \rightarrow \R^K$ for $K$ channels.
We define $S^2$ convolution as:
\begin{equation}
  \label{eq:s2-conv}
    f \conv \psi(R) = \int_{S^2} \sum_{k=1}^K f_k(x) \psi_k(R^{-1} x) dx
\end{equation}

It is important to note that in our definition, the result of convolution, $f \conv \psi$ is a function on the rotation group $\SO3$ and not a function on the sphere $S^2$.
As such, the convolution operation used in the following layers is slightly different; it is the $\SO3$ convolution:
\begin{equation}
  \label{eq:so3-conv}
  f \conv \psi(R) = \int_{\SO3} \sum_{k=1}^K f_k(R') \psi_k(R^{-1} R') dR'
\end{equation}

Our definition for spherical convolution differs from the more common one given by \cite{Driscoll1994}, where the result of convolution is a function on the sphere.
We found this definition too limiting, because it amounts to convolving with a filter that is rotationally symmetric around the north pole.
Other constructions that have as output a function on $S^2$ cannot be made to be equivariant.

Rotating a fixed filter over the sphere to detect patterns at every position and orientation on the sphere makes sense, because (by hypothesis) patterns appear with equal likelyhood in every position and orientation.
For this logic to hold also for the internal representations in the network, it is critical to show that the network layers are equivariant.

To this end we define the rotation operator $L_R f(x) = f(R^{-1} x)$ for functions on the sphere or $\SO3$.
It is easy to show, using definitions above and the invariance of the Haar measure \cite{Nachbin1965} that $(L_R f) * \psi = L_R (f * \psi)$ for both the $S^2$ and $\SO3$ convolution.
In other words, these operations are equivariant.

\section{Spherical FFT and Convolution Theorem}
\label{sec:fft}

It is well known that planar convolutions can be computed efficiently using the Fast Fourier Transform (FFT).
The Fourier theorem states that the Fourier transform of the convolution equals the element-wise product of the Fourier transforms, i.e. $\widehat{f \conv \psi} = \hat{f} \cdot \hat{\psi}$.
Since the FFT can be computed in $O(n \log n)$ time and the element-wise product has linear complexity, implementing the convolution using FFTs is asymptotically faster than the naive $O(n^2)$ spatial implementation.

For functions on the sphere and rotation group, there is an analogous transform, which we will refer to as the generalized Fourier transform (GFT) and a corresponding fast algorithm (GFFT).
This transform finds it roots in the representation theory of groups, but due to space contraints we will not go into details here and instead refer to interested reader to \citet{Sugiura1976, Folland1995}.

Abstractly, the GFT is nothing more than the projection of a function on a set of orthogonal basis functions called ``matrix element of irreducible unitary representations''.
For $\SO3$, these are the Wigner D-functions $D^l_{mn}(R)$ indexed by $l \geq 0$ and $-l \leq m, n \leq l$.
For $S^2$, these are the spherical harmonics\footnote{Technically, spherical harmonics are not matrix elements of irreducible representations, but $Y^l_m = D^l_{m0}|_{S^2}$} $Y^l_m(x)$ indexed by $l \geq 0$ and $-l \leq m \leq l$.

Denoting the manifold (e.g. $S^2$ or $\SO3$) by $X$ and the basis functions by $U^l$ (which is either vector-valued or matrix-valued), we can write the GFT of a function $f : X \rightarrow \R$ as
\begin{equation}
    \hat{f}^l = \int_{X} f(x) \overline{U^l(x)} dx
\end{equation}
This integral can be computed efficiently using a G-FFT algorithm \cite{Kostelec2007, Kostelec2008, Kunis2003, Driscoll1994, Maslen1998, Potts2009}.
For algorithmic details we refer the interested reader to \cite{Kostelec2007}.

The inverse $\SO3$ FT is defined as
\begin{equation}
  f(R) = \sum_{l=0}^b (2 l + 1) \sum_{m,n = -l}^l \hat{f}^l_{mn} U^l_{mn}(R)
\end{equation}
and similarly for $S^2$.
The maximum frequency $b$ is known as the bandwidth, and is related to the resolution of the spatial grid \cite{Kostelec2007}.

\begin{figure*}
  \centering
    \includegraphics[width=.9\textwidth]{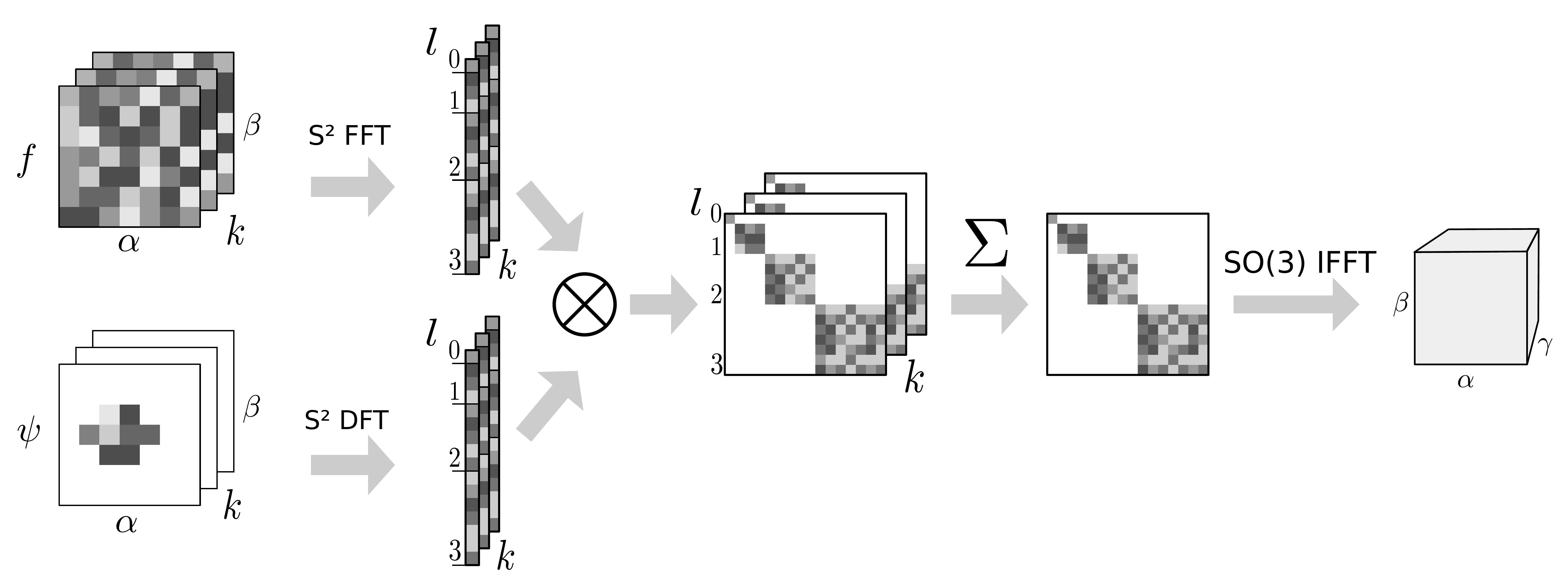}
  \caption{Spherical convolution in the spectrum. The signal $f$ and the locally-supported filter $\psi$ are Fourier transformed, tensored, summed over input channels, and finally inverse transformed. Note that because the filter is locally supported, it is faster to use a matrix multiplication (DFT) than an FFT algorithm for it. We parameterize the sphere using spherical coordinates $\alpha, \beta$, and $\SO3$ with ZYZ-Euler angles $\alpha, \beta, \gamma$.}
  \label{fig:spectral-conv}
\end{figure*}

Using the well-known (in fact, defining) property of the Wigner D-functions that $D^l(R)D^l(R') = D^l(RR')$, it can be shown that the $\SO3$ Fourier transform satisfies a convolution theorem: $\widehat{f \conv \psi} = \hat{f} \cdot \hat{\psi}$, where $\cdot$ denotes matrix multiplication of two block-diagonal matrices $\hat{f}$ and $\hat{\psi}$.
Similarly, using $Y(Rx) = D(R) Y(x)$ and $Y^l_m = D^l_{m0}|_{S^2}$, one can derive that $\widehat{f \conv \psi} = \hat{f} \cdot \hat{\psi}^\dagger$.
That is, the $\SO3$-FT of the $S^2$ convolution (as we have defined it) of two spherical signals can be computed by taking the outer product of the $S^2$-FTs of the signals.
This is shown in figure \ref{fig:spectral-conv}.

We were unable to find a reference for the latter version of the $S^2$ Fourier theorem, but it seems likely that it has been derived before.
The simplicity of the result shows that our definition of spherical convolution is natural.

\section{Spherical Convolutional Networks}
\label{sec:s2cnn}

A spherical ConvNet is constructed as follows.
The input signal $f : S^2 \rightarrow \R^{K_0}$ is convolved with a set of learnable filters $\psi^1_j, j=1, \ldots, K_1$ using the spherical convolution defined in eq. \ref{eq:s2-conv}.
This produces $K_1$ feature maps on $\SO3$.
The feature maps are composed with a non-linearity and then convolved again with a set of $\SO3$ filters $\psi^2_j$, etc.
At the end, we use a learned linear map and softmax nonlinearity to produce a distribution over classes.

Generally, we gradually reduce the resolution / bandwidth of the signal, while simultaneously increasing the number of channels.

More elaborate architectures such as residual networks with batch normalization can also be used.

\section{Experiments}
\label{sec:experiments}

\subsection{Numerical tests of Equivariance}

We have shown mathematically that the spherical convolution is equivariant, but this proof assumes that we are dealing with well-behaved (e.g. continuous, differentiable) functions.
In reality, we only have a set of samples $f(x^i)$ on some sampling set $\{x^i\}_i$.
It is therefore reasonable to ask to what degree the computation we actually perform is equivariant.
If there are severe artefacts that increase with network depth, we may find that equivariance is eventually lost and rotational weight sharing loses its effectiveness in a deep network.

\begin{figure*}[ht]
    \centering
    \input{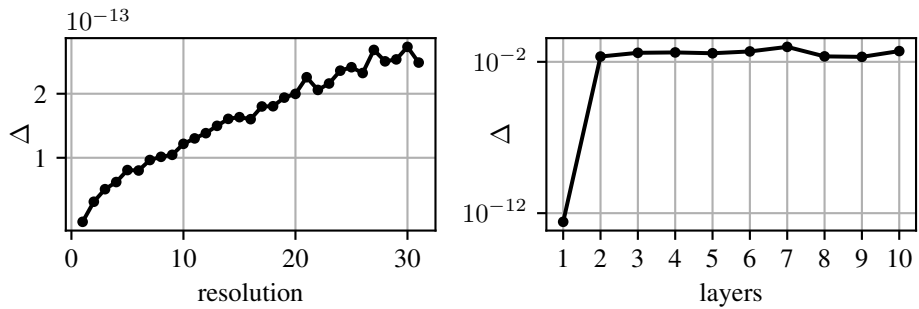}
    \caption{$\Delta$ as a function of the resolution or the number of layers (separated by ReLU activation functions). In both cases, we use $k=10$ channels. The error increases linearly with the resolution. The error increases sharply after the first ReLU layer, and then stays relatively constant.}
    \label{fig:equivariance}
\end{figure*}

To test the equivariance of the $\SO3$ convolution layer, we sample $n=500$ random rotations $R_i$ and $n$ feature maps $f_i$ with $K=10$ channels.
We then compute the average discrepancy $\Delta=\frac{1}{n}\sum_{i=1}^n \|L_{R_i} \Phi(f_i) - \Phi(L_{R_i} f_i)\|^2 / \|\Phi(f_i)\|^2$,
where $\Phi$ denotes an $\SO3$ convolution layer.
As shown in figure \ref{fig:equivariance}, the discrepancy remains manageable for a range of signal resolutions of interest.

In our second experiment we replace the group convolution layer with a ReLU network with $L=1, \ldots, 10$ layers.
The error increases rapidly to about $10^{-2}$ in the first two layers and then stays there.

\subsection{Spherical MNIST}

To find out if the inductive bias of a spherical CNN is indeed useful for classifying data with a rotationally invariant distribution, we constructed a spherical MNIST dataset.
The dataset is constructed by projecting digits onto the sphere, followed by an optional random rotation.

We use a conventional planar convolutional network as a baseline.
The network consists of two conv+relu layers, followed by a fully connected layer and a softmax.
We use $5 \times 5$ filters, $K=58, 114, 10$ channels and stride $3$ in both convolution layers.
Our spherical CNN has a similar architecture: S2Conv-relu-SO3Conv-relu-fc-softmax, with bandwidth $b=30,10,5$ and $k=100,200,10$.
Both models have about 165K parameters.

We evaluate both architectures in three regimes:
NR/NR, in which neither the train nor the test data is rotated, R/R in which both are randomly rotated, and NR/R, where only the test data is rotated.
Table \ref{tab:mnist-results} shows that the spherical CNN shows a slight decrease in performance for non-rotated data, but drastically outperforms the planar CNN on rotated data.
Most interestingly, the accuracy of the spherical CNN hardly drops in the NR / R regime, whereas the planar CNN deteriorates to chance level.

\begin{table}[h]
  \centering
  \begin{tabular}{c|ccc}
    & NR / NR & R / R & NR / R \\
    \hline
    planar & 0.99 & 0.45 & 0.09 \\
    spherical & 0.91 & 0.91 & 0.85 \\
    \hline
  \end{tabular}
  \caption{Test accuracy for the networks evaluated on the spherical MNIST dataset. Here R = rotated, NR = non-rotated and X / Y denotes, that the network was trained on X and evaluated on Y.}
  \label{tab:mnist-results}
\end{table}

\section{Conclusion}

In this paper we have presented spherical convolutional networks, which exploit prior knowledge about rotational symmetry of spherical signals in the same way a conventional convolutional network exploits translation symmetry of planar signals.
We have demonstrated mathematically as well as empirically that our spherical and $\SO3$ convolution layer is equivariant, so that it can be used effectively in deep networks.
Furthermore, we have developed an efficient convolution algorithm based on the generalized FFT.
Finally, we have demonstrated the effectiveness of the spherical CNN architecture for rotation-invariant classification of spherical signals.

In future work, we aim to apply the spherical CNN to problems in several important scientific problems such as molecular prediction problems \cite{Eickenberg} and global climate and metereological data.
In addition, we think spherical CNNs will be useful for analyzing visual data from omnidirectional cameras as well as 3D sensors.

\bibliographystyle{icml2017}

\end{document}